\def\year{2019}\relax
\documentclass[letterpaper]{article} 
\usepackage{aaai19}  
\usepackage{times}  
\usepackage{helvet}  
\usepackage{courier}  
\usepackage{url}  
\usepackage{graphicx}  
\usepackage{hyperref}
\usepackage{amsmath}
\usepackage{amsfonts}
\usepackage{amssymb}
\usepackage{natbib}
\usepackage{xcolor}
\usepackage{subfig}
\usepackage[ruled,vlined]{algorithm2e}

\newcommand{\TODO}[1]{}

\def\state{s}

\def\act{a}

\def\State{\mathcal{S}}
\def\Action{\mathcal{A}}

\def\Trans{T}

\def\Rew{R}
\def\rew{r}

\def\R{\mathbb{R}}
\def\E{\mathbb{E}}

\newcommand{\param}{\theta}
\newcommand{\policy}{\pi}

\newcommand{\edges}{E}

\newcommand{\Act}{\mathcal{A}}

\newcommand{\Q}{\qValue}

\newcommand{\fwcost}{F}
\newcommand{\fw}{\fwcost}
\newcommand{\qValue}{Q}

\newcommand{\epiT}{T}

\newcommand{\goal}{g}
\newcommand{\pos}{x}
\newcommand{\fwargs}[5]{\fw_{#4}^{#5}\left({#3}\middle|{#1}, {#2}\right)}
\newcommand{\Rgoal}{R_{\text{goal}}}

\title{Floyd-Warshall Reinforcement Learning:\\
  Learning from Past Experiences to Reach New Goals
  \thanks{Several weeks after submitting the work we found a work from
    \citet{kaelbling1993learning}, with similar main contribution is as our work.
    We have highlighted the minor differences in the related work
    section.}}

\author{Vikas Dhiman$^1$, Shurjo Banerjee$^1$, Jeffrey M. Siskind$^2$ and Jason J.
Corso$^1$\\
The University of Michigan$^1$\\
Purdue University$^2$}

\begin{document}

\maketitle
\begin{abstract}
Consider mutli-goal tasks that involve static environments and dynamic
goals. Examples of such tasks, such as goal-directed navigation
and pick-and-place in robotics, abound.
Two types of Reinforcement Learning (RL) algorithms are used for
such tasks: 
\emph{model-free} or \emph{model-based}. Each of these approaches has limitations.
Model-free RL struggles to transfer learned information when the goal location
changes, but achieves high asymptotic accuracy in single goal tasks. Model-based
RL can transfer learned information to new goal locations by retaining the
explicitly learned state-dynamics, but is limited by the fact that small errors in
modelling these dynamics accumulate over long-term planning.
In this work, we improve upon the limitations of model-free RL in
multi-goal domains. 
We do this by adapting the Floyd-Warshall algorithm for RL and call the
adaptation Floyd-Warshall RL (FWRL).
The proposed algorithm learns a goal-conditioned action-value function by 
constraining the value of the
optimal path between any two states to be greater than or equal to the value of
paths via intermediary states.
Experimentally, we show that FWRL is more sample-efficient and learns
higher reward strategies in multi-goal tasks as compared to Q-learning, model-based RL 
and other relevant baselines in a tabular domain.


\end{abstract}

\section{ Introduction}

Reinforcement learning (RL) allows for agents to learn complex behaviors
in a multitude of environments while requiring supervision only in the
form of reward signals. RL has had success in various domains ranging
from playing Atari games~\citep{MnKaSiNATURE2015} from purely visual
input and defeating world GO champions~\citep{gibney2016google}, to
applications in robotic navigation~\citep{mirowski2018learning} and
manipulation~\cite{levine2018learning}. In the realm of multi-goal tasks,
this work introduces Floyd-Warshall Reinforcement Learning (FWRL), a new
algorithm that facilitates the transfer of learned behaviours in
environments with dynamic goal locations.

There are two types of RL algorithms, \emph{model-based} and
\emph{model-free}, which differ in whether the state-transition function
is learned explicitly or implicitly~\citep{SuBaBOOK1998}.  In
\emph{model-based} RL, the dynamics that govern an environment's
transitions is explicitly modelled.  At any point in an episode, agents
use this model to predict future states and utilize this information to
maximize possible reward. This formulation is known to be
sample-efficient while normally not achieving high asymptotic
performance~\citep{pong2018temporal}.  In contrast, in \emph{model-free}
RL, algorithms such as policy gradients, actor-critic and Q-learning
directly learn the expected ``value'' for each state without explicitly
learning the environment-dynamics. This paradigm has been behind most of
the successes in such diverse applications
like Atari games, Go championships etc.

\begin{figure}%
\includegraphics[width=\columnwidth]{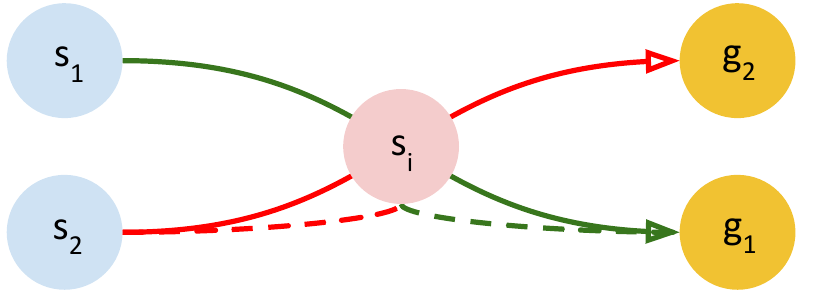}\\
\caption{The intuition driving Floyd-Warshall Reinforcement Learning.
Consider an agent experiencing traversals from $s_1$ to $g_1$ and then from
$s_2$ to $g_2$. Assume that there is at least one common state $s_i$ in
the two trajectories. The agent is then directed to traverse to $g_1$ from
$s_2$. The agent can \emph{exploit} previous trajectories to find a path to
perform this traversal (the dotted line), without requiring extensive
\emph{exploration}. However, standard Q-Learning cannot
make these generalizations. FWRL utilizes a triangle-inequality like constraint
(Eq~\eqref{eq:fwconstraint}) to combine trajectories from past experiences.
}
\label{fig:visual-abstract}%
\end{figure}

In multi-goal tasks that involve static environments and dynamic goals,
such as goal-directed navigation and pick-and-place in robotics,
model-free RL struggles to transfer learned behavior from one goal
location to another within the same
environment~\citep{dhiman2018critical,quillen2018deep}. This occurs because model-free
RL represents the environment as value functions, which conflate the
state-dynamics and reward distributions into a single representation.
On the other hand, model-based RL allows for the separation of
environment dynamics and reward, but can have lower asymptotic
performance due to the accumulation of small errors in the modelling function.

In this work we introduce Floyd-Warshall Reinforcement Learning, an
adaptation of the Floyd-Warshall shortest path
algorithm~\citep{floydwarshall1962}, for multi-goal tasks.
The Floyd-Warshall shortest path algorithm is itself a generalization of
Dijkstra's algorithm for multi-goal domains on graphs. The algorithm works
by learning a goal conditioned value function
~\citep{schaul2015universal}, called the Floyd-Warshall (FW) function,
that is defined to be the expected reward in going from a start
state-action pair $(\state, \act)$ to a given goal state $\state'$:
\begin{align}
\fwargs{\state}{\act}{\state'}{\policy}{} =
\E_{\policy}\left[ \sum_{t=0}^{t=k} \rew_t \middle\vert \state_0 = \state, \act_0 = \state, \state_k = \state' \right] .
\end{align}%
In order to learn the FW-function, we employ the following
triangular-inequality constraint for shortest paths, which is our main
contribution,
\begin{multline}
\fwargs{\state_i}{\act_i}{\state_j}{\policy^*_{\state_j}}{*}
 \ge 
  \fwargs{\state_i}{\act_i}{\state_k}{\policy^*_{\state_k}}{*}
  + \max_{\act_k}\fwargs{\state_k}{\act_k}{\state_j}{\policy^*_{\state_j}}{*}
  \\
  \forall \state_k \in \State.
  \label{eq:fwconstraint}
\end{multline}%
Section~\ref{sec:method} describes the terminology, equations and the method in more detail.

This constraint allows FWRL to remember paths even if they do not lead
to the goal location during particular episodes. The motivation is
similar to the Hindsight Experience Replay
(HER)~\citep{anderson2017vision}, where agents learn by re-imagining the
final states of past failed experiences as succesful. Our method allows
us to utilize hindsight experience in more a fine-grained manner because
we consider all states as potential goals. Other works using a
goal-conditioned value function for multi-goal tasks include Universal
Value Function Approximators (UVFA)~\citep{schaul2015universal} and
Temporal Difference Models (TDM)~\citep{pong2018temporal}. UVFA
introduces the use of goal-conditioned value functions and a
factorization approach for faster learning of neural networks that
depend upon goals from the state space.  TDM combines model-based and
model-free algorithms by modeling the learning as a constrained
objective function using a horizon dependent goal-conditioned value
function. In contrast to these works, we present an alternative
mechanism for learning these functions that is horizon independent.  

Experimentally, FWRL is shown to be more sample-efficient and achieve higher
reward standard Q-Learning and model-based methods in a tabular setting. FWRL is
found to achieve two times median rewards than the next best baseline.



\section{Related work}

\paragraph{Goal-conditioned value functions}
Multi-goal reinforcement learning has gained attention lately with
important work like UVFAs~\citep{schaul2015universal},
HER~\citep{andrychowicz2016learning} and TDM~\citep{pong2018temporal}
making progress towards learning goal-conditioned value functions for
multi-goal tasks.  Introduced by ~\citep{schaul2015universal}, universal
goal-conditioned value functions (UVFs), $V(s,g)$, measures the utility
function of achieving any goal from any state in an environment. In this
work, UVFs are learned using traditional Q-learning based approaches
coupled with matrix factorization based methods for faster learning.
In Hindsight Experience Replay (HER), \citet{andrychowicz2016learning}
learn UVFs from previous episodes
accounting for when the goal location has not been achieved.  Their
method works by utilizing failed past experiences and re-imagining 
the last states of these episodes as goal states.
\citet{pong2018temporal} propose Temporal Difference Models (TDM) that estimate
goal directed horizon dependent value functions, $Q(s, a, g, \tau)$. There work is
limited because they assume that rewards are available densely in the form of
some distance like measure from the goal.
In contrast to all the above works our contribution is a novel algorithm for
learning UVFs via leveraging a triangular-inequality like constraint within the
space of these functions.

\paragraph{Goal directed visual navigation}
There has been considerable interest in using Deep Reinforcement
Learning (DRL) algorithms for the goal-driven visual navigation of
robots~\citep{mirowski2016learning,MiPaViICLR2017,dhiman2018critical,gupta2017cognitive,savinov2018semi}.
\cite{mirowski2016learning} demonstrate that a DRL algorithm called
Asynchronous Advantage Actor Critic (A3C) can learn to find a goal in 3D
navigation simulators, using only a front facing first person view as
input, while \cite{MiPaViICLR2017} demonstrate goal directed navigation
in Google's street view graph. \citet{gupta2017cognitive} addresses goal
directed mapping but their method depends upon
navigation-specific occupancy grids which limit the method's
generalizability to mutli-goal tasks. Moving the successes of these
works from simulations to the real world is an active area of research 
because of the high sample complexity of
model-free RL algorithms \citep{zhu2017target,anderson2018vision} . \citet{dhiman2018critical} empirically evaluate
\cite{mirowski2016learning}'s approach and show that when goal locations
are dynamic, the path chosen to reach the goal are often far from
optimal.  In contrast to our method, these works focus on the
navigation domain and employ domain specific auxiliary rewards and
data-structures making them less generalizable to other multi-goal
tasks.

\paragraph{Binary goal tasks}
In \cite{OhChSiICML2016}, agent's DRL-driven navigation is complimented
with external memory modules indexed in time.
In a similar vein, \citet{parisotto2017neural} experiment with indexing
memory modules with the agent's spatial coordinates. Both works 
consider goal specification as a binary signal hidden somewhere in the
observation space. In such problems the challenge is to learn the goal
specification protocol while avoiding the wrong goal and exploring for the right
one. These problems domains do not evaluate the ability of agents to find the
shortest path to the goal.

\paragraph{Model-based RL}
Model-based approaches are known to have lower asymptotic performance then
model-free approaches~\cite{pong2018temporal}. In many tasks, like moving a
glass of water, it is easier to model the expected rewards rather than state
dynamics. This is because inaccurate modeling of the water-surface-dynamics
won't affect the rewards as long the agent does not spill large amounts of
water. More recently, model-based algorithms have shown more promise by
explicitly modelling uncertainty~\cite{lakshminarayanan2017simple,
  kurutach2018model,zhang2018solar}.
However, more work is needed for model-free algoritms to be replaced by
model-based ones. 

Our work is most similar to \citet{kaelbling1993learning} who present a
reinforcement learning method to reach goals through shortest path using the
relaxation constraint from Floyd-Warshall. In contrast, we do not restrict our
formulation to step lengths based shortest path, instead we formulate the
problem in terms of rewards, like typically done in reinforcement learning, to achieve
maximum reward path.

%

%
%

\section{Background}

We present a short review of the background material that our work depends upon.

\subsection{Dijkstra}
\newcommand{\vertices}{\State}
\newcommand{\edge}{\rew}
\newcommand{\fwds}{D}
\newcommand{\dds}{D}
Dijkstra~\citep{dijkstra1959note} is a shortest path finding algorithm from a
given vertex in the graph. Consider a weighted graph $G = (\vertices,
\edges)$, with $\vertices$ as the vertices and $\edges$ as the edges. Dijkstra
algorithms works by maintaining a data-structure $\dds : \vertices \rightarrow
\R$, that represents the shortest path length from the source. The data
structure $\dds$ is initialized with zero at start location $\dds[\state_0]
\leftarrow 0$ and a high value everywhere else $\dds[i] \leftarrow \infty \, \forall
i \in \vertices$. The algorithm then sequentially updates $\dds$ by
\begin{align}
  \dds[j] \leftarrow \min\{\dds[j], \edge_{(i, j)} + \dds[i] \} \, \forall (i, j) \in \edges ,
  \label{eq:dijkstra}
\end{align}%
where $\edge_{(i, j)}$ is the edge-weight for directed edge $(i, j) \in \edges$.
The shortest path $(\state_0, \state_{1}, \dots)$ starting from vertex
$\state_0$ can be read from $\dds$ via $\state_{t+1} = \arg \min_{i \in
\text{Nbr}(\state_t)} \dds[i]$ where $\text{Nbr}(\state_t) = \{ i | (i,
\state_t) \in \edges \} $ denotes the neighborhood of $\state_t$. With a
carefully chosen data-structure and traversal order, the Dijkstra Algorithm can
be made to run in $O(|\vertices|\log|\vertices|)$.

\subsection{Q-Learning}
Q-learning~\citep{watkins1992qlearning} is a reinforcement learning (RL)
algorithm that allows agent to explore environment and simultaneously
compute maximum reward paths.

An RL problem is formalized as an Markov Decision Process (MDP). A MDP is
defined by a four tuple $(\State, \Action, \Trans, \Rew)$, where $\State$ is the
state space, $\Action$ is the action space, $\Trans : \State \times \Action
\rightarrow \State$ is the system dynamics and $\Rew : \State
\rightarrow \R $ is the reward yielded on a execution of an action.
The objective of a typical RL problem is to maximize the expected cumulative
reward over time, called the returns  $ R_t = \sum_{t^{'}=t}^{T} \rew_{t^{'}}$.

Q-learning works by maintaining an action-value function $\Q : \State \times
\Action \rightarrow \R$ which is defined as the expected return
$\Q_\policy(\state_t, \act_t) = \E_\policy[R_t]$ from a given state-action pair.
The Q-learning algorithm works by updating the $\Q$-function using the Bellman
equation for every transition from $\state$ to $\state'$ on action $\act$
yielding reward $\rew$, 
\begin{align}
  \Q^*(\state, \act) &= \E_{\policy}\left[
                       \rew + \max_{\act^{'}} \Q^*(\state^{'}, \act^{'})
                       \middle| \state, \act \right] \, .
    \label{eq:q-learn-bellman}
\end{align}%

\subsection{Floyd-Warshall}

The Floyd-Warshall algorithm~\citep{floydwarshall1962} is a shortest path finding
algorithm from any vertex to any other vertex in a graph.
Similar to Dijkstra's algorithm, the Floyd-Warshall algorithm
finds the shortest path by keeping maintaining a shortest distance
data-structure $\fwds : \vertices \times \vertices \rightarrow \R$. between any
two pair of vertices $i, j \in \vertices$.
The data-structure $\fwds$ is initialized with edges weights
$\fwds[i, j] \leftarrow \edge_{(i, j)} \, \forall (i, j) \in \edges$
and the uninitialized edges are assigned a high value,
$\fwds[i, j] \leftarrow \infty \, \forall i, j \in \vertices$.
The algorithm works by sequentially observing all the nodes in the graph and
updating $\fwds$ with the shortest explored path known so far:
\begin{align}
  \fwds[i, j] \leftarrow \min\{ \fwds[i, j], \fwds[i, k] + \fwds[k, j] \} \quad
  \forall i, j, k \in \vertices \, .
\end{align}%

The update equation in the algorithm depends upon the triangular inequality for
shortest paths distances ($\fwds[i, j] \le \fwds[i, k] + \fwds[k, j]$) and hence
works only in the absence of negative cycles in the graph. Fortunately, many
practical problems can be formulated such that negative cycles
do not occur. The Floyd-Warshall algorithm runs in $O(|\vertices|^3)$
and is suitable for dense graphs. There also exists extensions of the algorithm
like Johnson's algorithm~\citep{johnson1977efficient} that run in
$O(|\vertices|^2\log|\vertices| + |\vertices||\edges|)$ while working on
the same principle.

From the parallels between Eq.~\eqref{eq:dijkstra} and
Eq.~\eqref{eq:q-learn-bellman}, Q-learning can be seen as a generalization
Dijkstra's algorithm. Both the algorithms work by taking one step minimum (or
maximum) over the neighboring state. Unlike Dijkstra, in Q-learning one has to
compute an additional maximum over actions. This is because in an MDP, the agent
cannot directly choose the next state to be in. Instead, it chooses an action
that leads it to the next state based on transition probabilities. Moreover,
Q-learning has to explore the state space before it can exploit the learned
information to find most-rewarding path. With these parallels in mind, we
generalize the Floyd-Warshall algorithm to work on an MDP and call it
Floyd-Warshall Reinforcement Learning.

\section{Problem definition}

Consider an agent interacting with an environment, $\varepsilon$. At
every time step, $t$, the agent takes an action $\act_t \in \Act$, observes a
state, $\state_t \in \State$ and a reward $\rew_t \in [-\Rgoal, \Rgoal]$.
A goal state, $\goal \in \State$, is provided to the agent and it receives
$\Rgoal$--the highest reward in the environment--on reaching it
with respect to some threshold $\|\state -\goal\| < \delta_{\text{goal-thresh}}$
An episode is defined as of a fixed number of time steps, $T$. For
every episode, a new goal state is provided to the agent as input. If the agent
reaches the goal state before the episode ends, the agent is 
re-spawned at a new random location within the environment while the goal state
remains unchanged for the episode.
The agent's objective is to find the sequence of actions to take that maximizes
the total reward per episode. 


%
%

%

Model-free Reinforcement learning methods assume that the rewards are
being sampled from the a static reward function.  In a problem where the
goal location changes it becomes hard to transfer the learned
value-function or action-value function to the changed location.  One
alternative is to concatenate the goal location the state, making the
new state space $[\state_t, \goal]^\top \in \State^2$ larger.  This
method is wasteful in computation and more importantly in sample
complexity \cite{schaul2015universal}.

\section{Method}
\label{sec:method}
We present a model-free reinforcement learning method that transfers
learned behavior when goal locations are dynamic. We call this algorithm
Floyd-Warshall Reinforcement Learning, because of its similarity to
Floyd-Warshall algorithm~\cite{floydwarshall1962}:
a shortest-path planning algorithm on graphs. Similar
to universal value function~\cite{schaul2015universal}, we define the Floyd-Warshall
(FW) function as the expected cumulative reward on going from a start
state to an
end state within an episode:
\begin{align}
\fwargs{\state}{\act}{\state'}{\policy}{} =
\E_{\policy}\left[ \sum_{t=0}^{t=k} \rew_t \middle\vert \state_0 = \state, \act_0 = \state, \state_k = \state' \right] ,
\end{align}%
where $\policy$ is the
stochastic policy being followed.
Note that we do not use discounted rewards, instead assuming that episodes are
of finite time length. In keeping with the Floyd-Warshall shortest path
algorithm, we assume that there are no positive reward
cycles in the environment.

Note that the FW-function is closely related to the Q-function,
\begin{align}
  \Q_\policy(\state, \act) = \sum_{\state'} P_\policy(\state' | \state, \act) \fwargs{\state}{\act}{\state'}{\policy}{},
\end{align}%
where $P_\policy(\state' | \state, \act)$ is the probability of the agent
arriving at $\state'$ within the episode. We define the optimal FW-function as
the maximum expected value that a path between a start state and a goal state can
yield,
\begin{align}
\fwargs{\state}{\act}{\state'}{\policy^*_{\state'}}{*} =
\max_{\policy_{\state'}}  \fwargs{\state}{\act}{\state'}{\policy_{\state'}}{}
\end{align}%
where $\policy^*_{\state'}$ is the
optimal policy towards the goal state $\state'$. Once the
FW-function approximation is learned, the optimal policy can be computed from
FW-function similar to the Q-learning algorithm, $\policy^*_{\state'}(\state) =
\arg \max_{\act} \fwargs{\state}{\act}{\state'}{}{*}$

When the policy is optimal, the Floyd-Warshall function must satisfy the
constraint
\begin{multline}
\fwargs{\state_i}{\act_i}{\state_j}{\policy^*_{\state_j}}{*}
 \ge 
  \fwargs{\state_i}{\act_i}{\state_k}{\policy^*_{\state_k}}{*}
  + \max_{\act_k}\fwargs{\state_k}{\act_k}{\state_j}{\policy^*_{\state_j}}{*}
  \\
  \forall \state_k \in \State.
\end{multline}%
In other words, the value for the optimal path from given start state to a given
goal state should be greater than or equal to value of path via any intermediate
state.
This triangular-inequality like constraint is the main contribution of our work.
To the best of our knowledge it has not been employed in any previous works
utilizing goal-conditioned value functions.

Aside from the dynamic goal locations, we assume environments underlying reward
distributions to be static. We also assume that the goal reward is the
highest reward in the reward space.
The pseudo-code for the algorithm in shown Alg~\ref{alg:floyd-warshall-small}.

\begin{algorithm}
  \tcc{By default all states are unreachable}
  Initialize $\fwargs{\state_i}{\act_i ; \param_{\fwcost}}{\state_j}{}{} \leftarrow -\infty$ \;

  Define $\policy^*(\state_t, \state_g, \fw) = \arg \max_{\act}
  \fwargs{\state_t}{\act; \param_{\fwcost}}{\state_g}{}{}$ \;
  Input $\state_g$ \;
  Set $t \leftarrow 0$\;
  Observe $\state_t$ \;
  \For{$t \leftarrow 1$ \KwTo $\epiT$}{
    Take action $\act_{t} \sim \epsilon\text{-greedy}(\policy^*(\state_{t}, \state_g, \fw))$ \;
    Observe $\state_{t+1}$, $\rew_t$ \;
    \If{$\rew_t >= \Rgoal$}{
      \tcc{Do not update the value function with goal reward}
      continue\;
    }
    $\fwargs{\state_{t}}{\act_{t}}{\state_{t+1}}{}{} \leftarrow \rew_t$ \;
    \For{$\state_k \in \State, \act_k \in \Act, \state_l \in \State$}{
      $\fwargs{\state_k}{\act_k}{\state_l}{}{} \leftarrow
        \max \{
        \fwargs{\state_k}{\act_k}{\state_l}{}{},
        \fwargs{\state_k}{\act_k}{\state_t}{}{}
        + \max_{\act_p \in \Act} \fwargs{\state_t}{\act_p}{\state_l}{}{}
        \}$
        \;
    }
  }
  \KwResult{$\policy^*(\state_k, \state_g, \fwcost)$}
  \caption{\small Floyd-Warshall Reinforcement Learning (Tabular setting)}
  \label{alg:floyd-warshall-small}
\end{algorithm}

%

\section{Experiments}
\label{sec:experiments}

Experiments are conducted in a grid-world like setup as displayed in
figure \ref{fig:four-room-grid-world}. The agent can occupy any one of
the white blank squares. The agent's observations is the numbered
location of each square i.e. each squares x,y coordinate with the origin
being the top left corner of the environment, $s_t = (x, y)$. Agents
can act by moving in four  directions, $\{up, down, left, right\}$.
Experiments are conducted in a tabular domain showcasing results
that are intuitive to understand while still displaying large
performance gaps between FWRL and standard baselines methods. 

\section{Environment}
We use two types of grid world environments in the experiments, a four room grid
world and a windy version of the four room grid world.

\subsubsection{Four room grid world}
\begin{figure}[h!]%
\includegraphics[width=0.48\columnwidth]{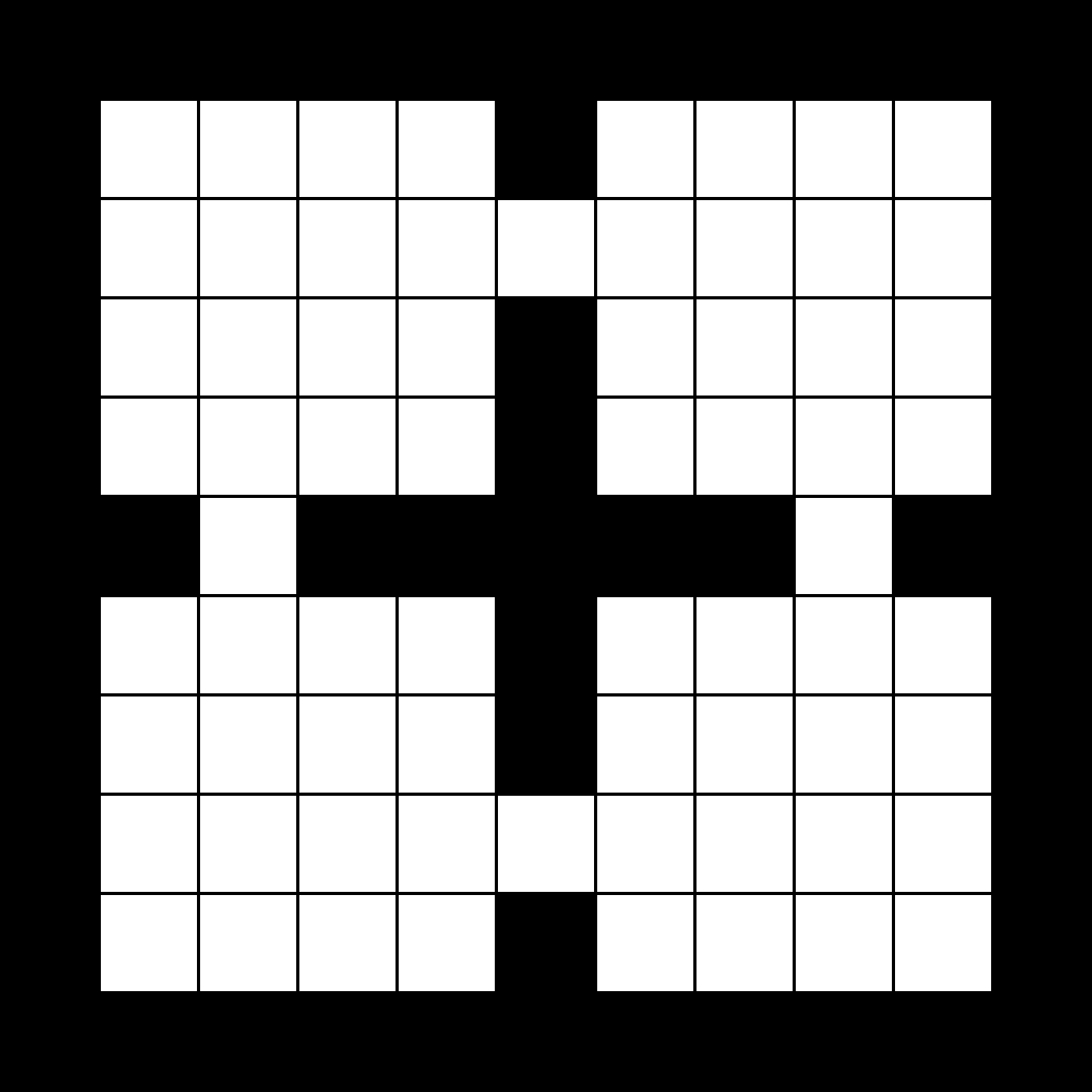}
\hfill
\includegraphics[width=0.48\columnwidth]{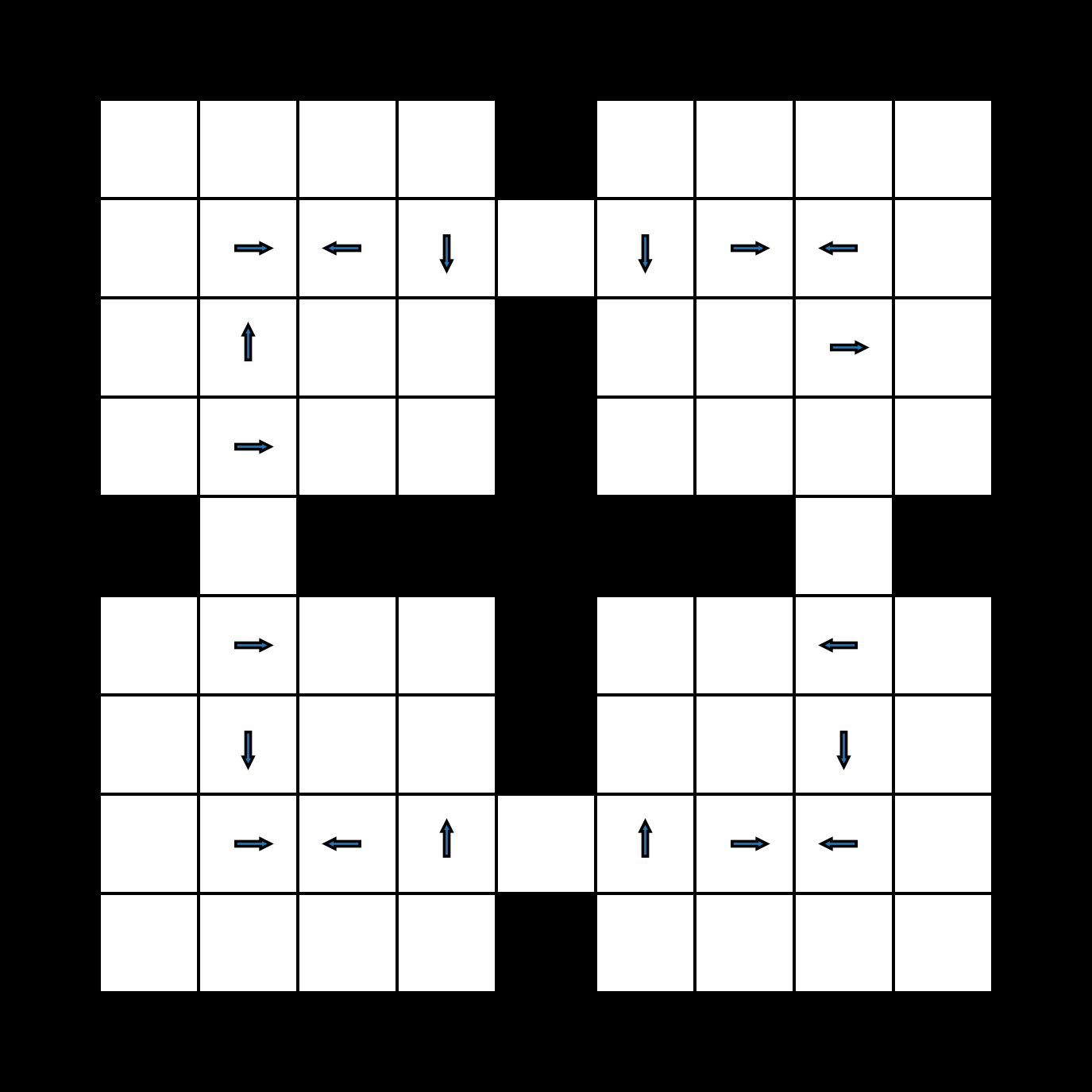}%
\caption{Left: Four room grid world. Right: Four room windy grid world with wind direction shown by arrows. The windy pushes the agent in the direction of wind with 0.25 probability irrespective of the action taken.}
\label{fig:four-room-grid-world}%
\end{figure}

Four room grid world is a grid world with four rooms connected to each
other as shown in Figure~\ref{fig:four-room-grid-world}. This example is
chosen due to it's intentional difficulty for random exploration based
agents. Since the exit points, are narrow, random agents tend to get
stuck in individual rooms. 

\subsubsection{Four room windy world}
In four room windy world, the previous setup is augmented in some cells
with \emph{wind}. This wind, indicated by arrows, increases the
probability of the agent randomly going in the direction of the arrow by
0.25.  Conceived in~\citet{SuBaBOOK1998}, this setup increases the
dependence of the dynamics model upon environment conditions. 

\subsection{Metrics}
The metrics used to quantify and compare agent performance across
FWRL to baseline methods are described here.

\subsubsection{Reward}
As in typical in reinforcement learning, the reward earned per episode by the
agent is treated as a metric of success.


\subsubsection{Distance-Inefficiency} The distance-inefficiency~\citep{dhiman2018critical} is the ratio of the
      distances travelled by the agent during an episode to the sum of the shortest
      path to the goal at every point of initialization. Mathematically it is defined
      as:
		\begin{align}
			\text{Dist-ineff.} &=
			\frac{ \sum_{i=1}^{N-1} \sum_{k=\tau_i + 1}^{\tau_{i+1} - 1} \|\pos_{k+1} - \pos_{j}\| }
			{ \sum_{i=1}^{N-1} \delta(\pos_{\tau_i + 1}, \pos_g) } ,
		\end{align}%
		where $\delta(\pos_{\tau_i +1}, x_g)$ denotes the shortest path
		distance between spawn location $\pos_{\tau_i+1}$ and goal location
        $\pos_g$. The numerator is the total distance covered by the agent while
        skipping the jumps where the agent gets re-spawned after reaching the
        goal location. The denominator is the total shortest distance during the
        episodes.

\subsection{Baselines}
We compare our FWRL against three baselines: two versions of
Q-Learning~\cite{watkins1992qlearning} and Model based RL (MBRL).

\subsubsection{Q-Learning: QL and QLCAT}
We implement two versions of Q-learning. In the first version, called QL, we
reset the Q-function after every episode. This version of Q-Learning does not
use the prior knowledge of the goal state, but depends upon the goal reward to
build a new Q-function in every episode. In the second version, called QLCAT,
we concatenate the state with the goal location and retain the learned Q-function
function across episodes. 

\subsection{Model-based RL}
We implement a simple version of tabular model-based RL where we maintain a
transition count data-structure $T(\state' | \state, \act)$.
This allows to compute a frequentist estimate of dynamics model.
We also keep a tabular record of the reward from each state-action pair
$R(\state, \act)$. The dynamics model is then used to find the
maximum-reward-path to the goal state.

\section{Results}
\subsubsection{Quantitative Results}
\begin{figure}%
  \includegraphics[width=\columnwidth]{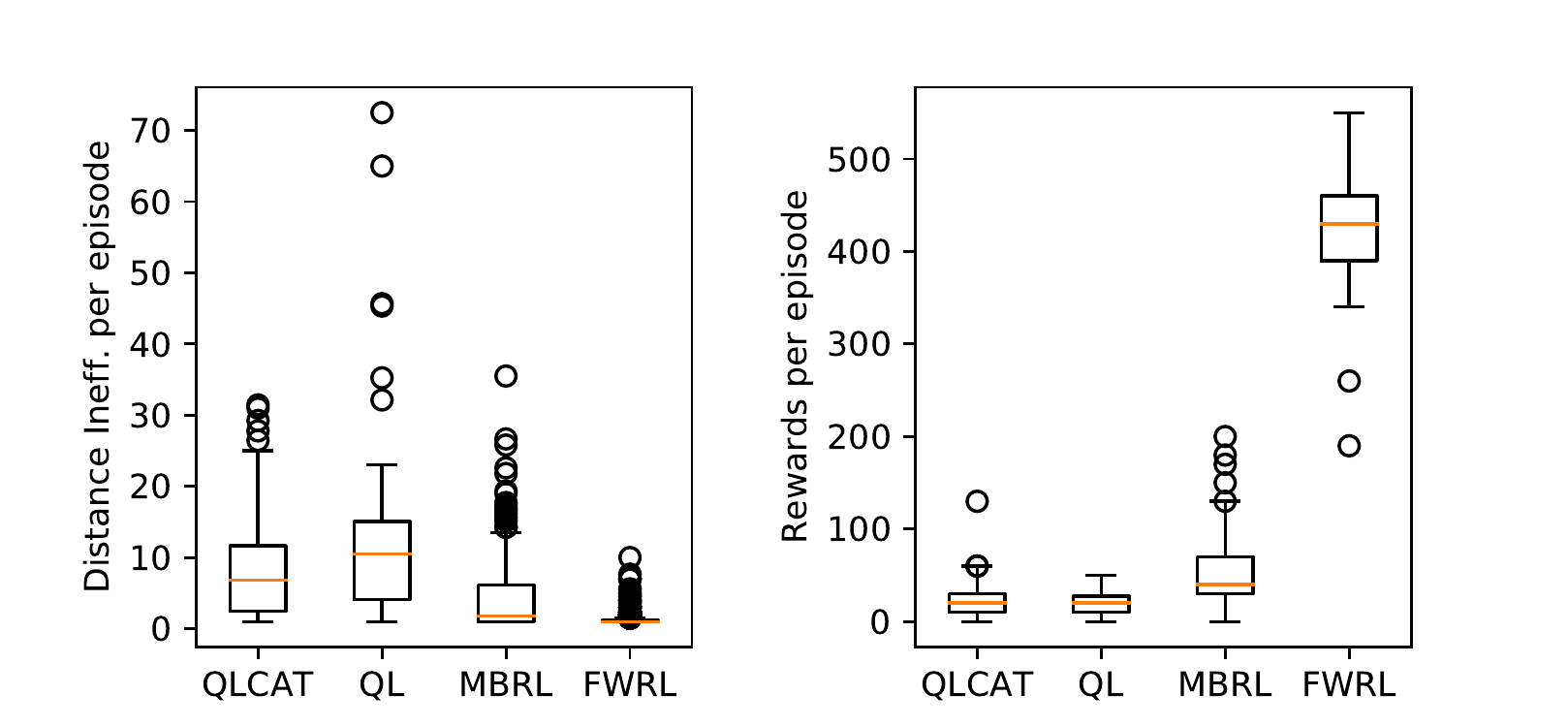}{a}
  \caption{Results on grid world. FWRL beats other baselines
    consistently. Lower is better for Distance-Inefficiency. Higher
    is better for reward per episode. }
  \label{fig:ql-fw-grid-world-results}%
\end{figure}

We evaluate Q-learning Concatenated (QLCAT), Q-learning (QL), model-based RL
(MBRL) and Floyd-Warshall Reinforcement Learning (FWRL) on two metrics in two
different environments. The two metrics we use are Distance inefficiency and
average reward per episode. The baselines and metrics are explained in the
experiment section (\ref{sec:experiments}). The results are shown in
Figure~\ref{fig:ql-fw-grid-world-results}-\ref{fig:ql-fw-windy-world-rewards} 
We find that FWRL consistently achieves higher rewards, and lower distance
inefficiency than the baselines. The reward curves also show that FWRL learns to
find the higher reward much faster than the baselines, demonstrating higher
sample efficiency.

\begin{figure}
  \includegraphics[width=\columnwidth]{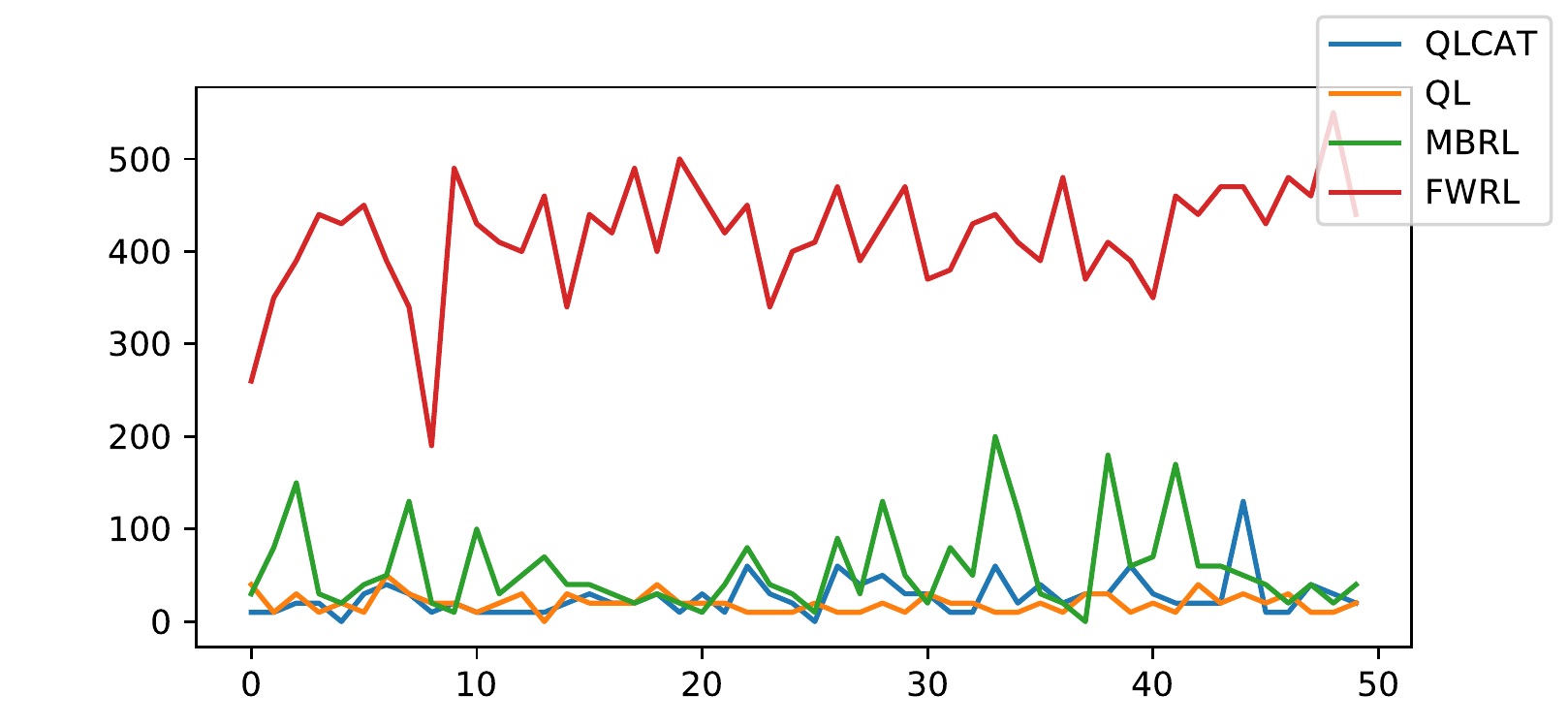}{b}
  \caption{Reward curves on grid world. FWRL reward climbs much
    faster than all other baselines showcasing the improved \emph{sample
      efficiency} of the algorithm.}
  \label{fig:ql-fw-grid-world-reward-curves}%
\end{figure}

\begin{figure}
  \includegraphics[width=\columnwidth]{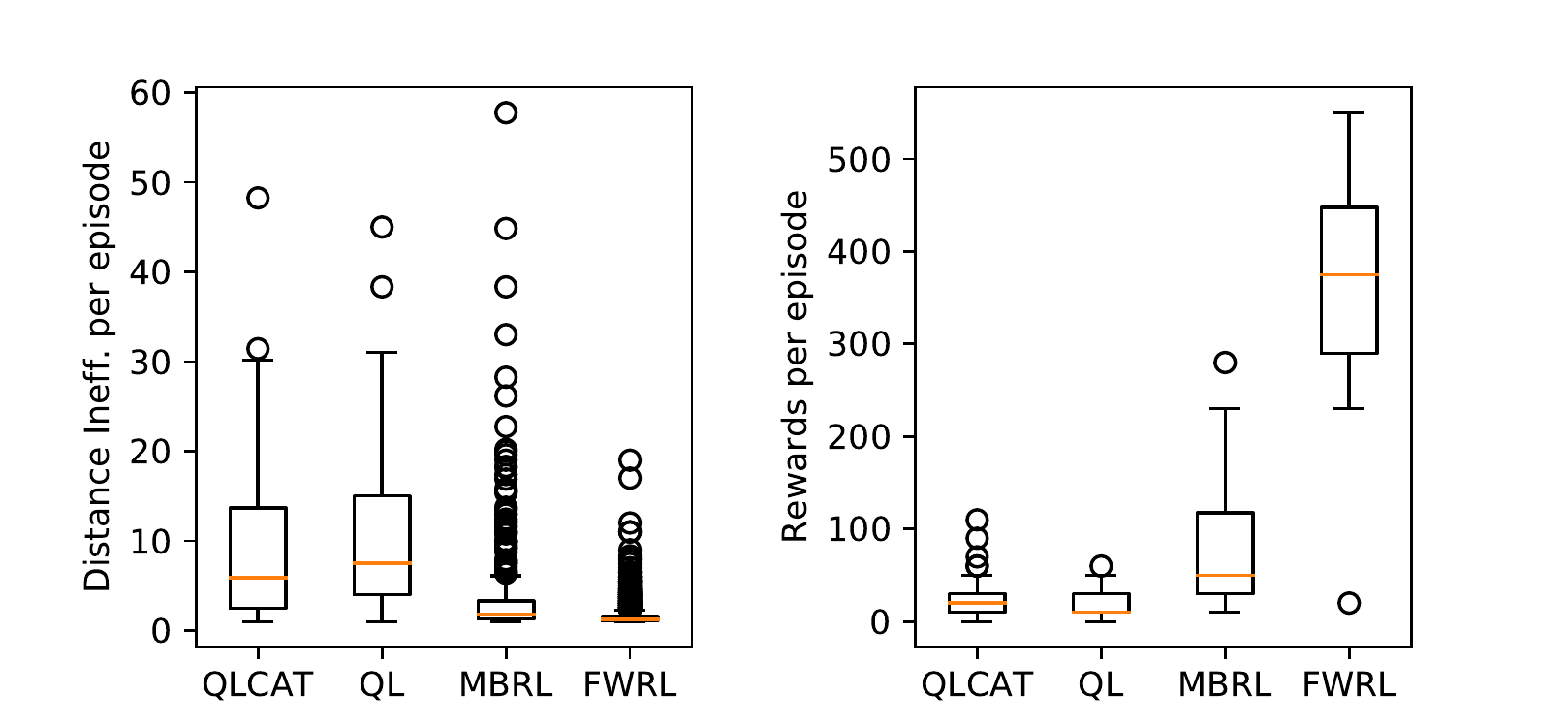}
  \caption{Results on windy world. FWRL beats other baselines
    consistently. Lower is better for Distance-Inefficiency. Higher
    is better for reward per episode. }
  \label{fig:ql-fw-windy-world-results}%
\end{figure}

\begin{figure}
  \includegraphics[width=\columnwidth]{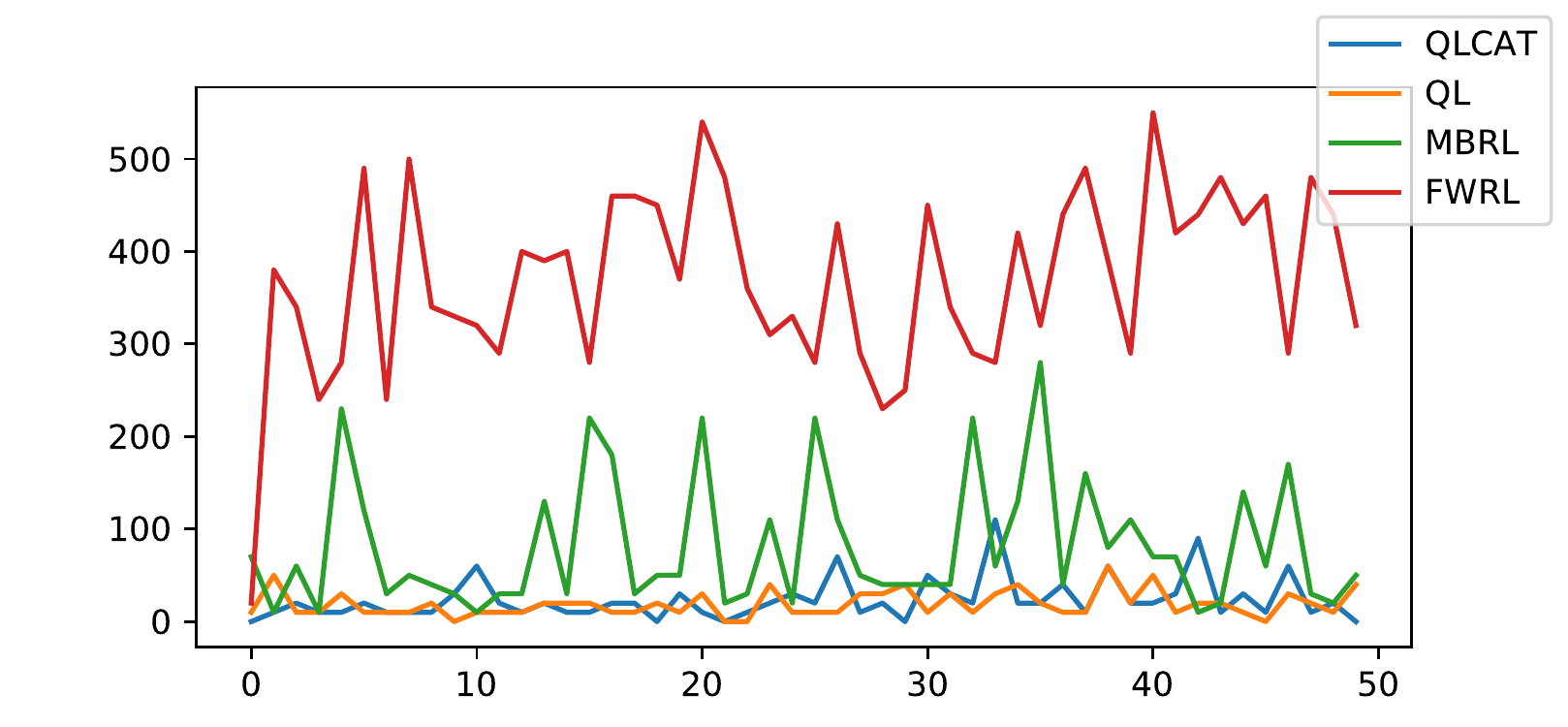}
  \caption{Reward curves on windy world. FWRL reward climbs much
    faster than all other baselines showcasing the improved \emph{sample
      efficiency} of the algorithm.}
  \label{fig:ql-fw-windy-world-reward-curves}%
\end{figure}

\subsubsection{Qualitative Results}
To demonstrate the claim made in Fig~\ref{fig:visual-abstract}, we train QLCAT
and FWRL for two episodes each with start and goal locations such that the path
meets in the center. Unlike the quantitative section, in this experiment the
episode ends when the agent reaches the goal.
After two episodes of training, in which both FWRL (Floyd-Warshall RL) and QLCAT
(Q-Learning with goal concatenated to the state) reach the desired
goals via exploration, we put the algorithms to test.
For the test episode, the goal is chosen from first training episode but start
location is chosen from second training episode.
During the test episode,
we turn off $\epsilon$-greedy exploration and follow the learned policy
greedily.

\begin{figure}
  \includegraphics[width=\columnwidth]{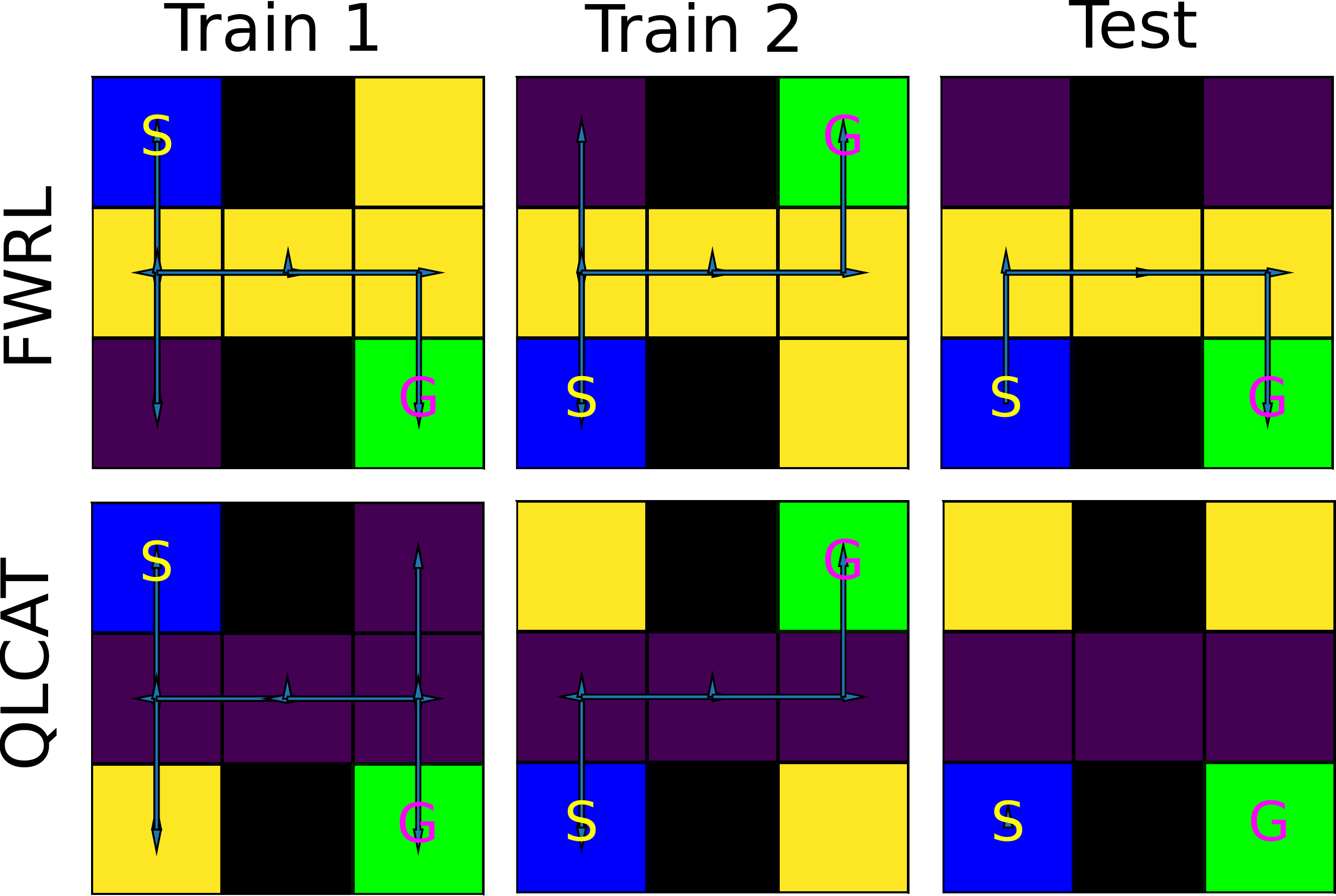}
  \caption{Qualitative results: Visualization of value-function in H-Maze. Green
box with G represents goal location, Blue box with S represents start location.
The obstacles are shown black. The trajectories are shown by arrows. The color
of remaining boxes show the expected value of each state computed by the
corresponding algorithm. Each row different algorithm, while columns show
different episodes. We find that QLCAT is unable to learn a new trajectory
when given an unseen combination of start and end location. FWRL (ours) finds the
shortest path easily in such a test case. }
\label{fig:qualitative-results}
\end{figure}

We find that QLCAT decides to repeat the action that pushes it into the wall,
therefore is unable to move. However, FWRL reaches the goal using the shortest
path. The trajectories and value functions are visualized in Figure~\ref{fig:qualitative-results}.

\section{Conclusion}
Floyd-Warshall Reinforcement Learning (FWRL) allows us to learn a goal
conditioned action-value function which is invariant to the change in
goal location as compared to the action-value functions typically used
in typical Q-learning.  This allows FWRL to transfer learned behaviors
about the environment when the goal location changes. Many
goal-conditioned tasks like navigation and robotic pick and place can
benefit from this framework.

\subsubsection*{Acknowledgements}
This work was supported, in part, by the US National Science Foundation under
Grants 1522954-IIS and 1734938-IIS, by the Intelligence Advanced Research
Projects Activity (IARPA) via Department of Interior/Interior Business Center
(DOI/IBC) contract number D17PC00341, and by Siemens Corporation, Corporate
Technology.
Any opinions, findings, views, and conclusions or recommendations expressed in
this material are those of the authors and do not necessarily reflect the
views, official policies, or endorsements, either expressed or implied, of the
sponsors.
The U.S. Government is authorized to reproduce and distribute reprints for
Government purposes, notwithstanding any copyright notation herein.

\def\localbib{/home/dhiman/wrk/group-bib/shared}
\IfFileExists{\localbib.bib}{
\bibliography{\localbib,main,main_filtered}}{
\bibliography{main,main_filtered}}

\begin{thebibliography}{}

\bibitem[\protect\citeauthoryear{Anderson \bgroup et al\mbox.\egroup
  }{2017}]{anderson2017vision}
Anderson, P.; Wu, Q.; Teney, D.; Bruce, J.; Johnson, M.; S{\"u}nderhauf, N.;
  Reid, I.; Gould, S.; and Hengel, A. v.~d.
\newblock 2017.
\newblock Vision-and-language navigation: Interpreting visually-grounded
  navigation instructions in real environments.
\newblock {\em arXiv preprint arXiv:1711.07280}.

\bibitem[\protect\citeauthoryear{Anderson \bgroup et al\mbox.\egroup
  }{2018}]{anderson2018vision}
Anderson, P.; Wu, Q.; Teney, D.; Bruce, J.; Johnson, M.; S{\"u}nderhauf, N.;
  Reid, I.; Gould, S.; and van~den Hengel, A.
\newblock 2018.
\newblock Vision-and-language navigation: Interpreting visually-grounded
  navigation instructions in real environments.
\newblock In {\em Proceedings of the IEEE Conference on Computer Vision and
  Pattern Recognition (CVPR)}, volume~2.

\bibitem[\protect\citeauthoryear{Andrychowicz \bgroup et al\mbox.\egroup
  }{2016}]{andrychowicz2016learning}
Andrychowicz, M.; Denil, M.; Gomez, S.; Hoffman, M.~W.; Pfau, D.; Schaul, T.;
  Shillingford, B.; and De~Freitas, N.
\newblock 2016.
\newblock Learning to learn by gradient descent by gradient descent.
\newblock In {\em Advances in Neural Information Processing Systems},
  3981--3989.

\bibitem[\protect\citeauthoryear{Dhiman \bgroup et al\mbox.\egroup
  }{2018}]{dhiman2018critical}
Dhiman, V.; Banerjee, S.; Griffin, B.; Siskind, J.~M.; and Corso, J.~J.
\newblock 2018.
\newblock A critical investigation of deep reinforcement learning for
  navigation.
\newblock {\em arXiv preprint arXiv:1802.02274}.

\bibitem[\protect\citeauthoryear{Dijkstra}{1959}]{dijkstra1959note}
Dijkstra, E.~W.
\newblock 1959.
\newblock A note on two problems in connexion with graphs.
\newblock {\em Numerische mathematik} 1(1):269--271.

\bibitem[\protect\citeauthoryear{Floyd}{1962}]{floydwarshall1962}
Floyd, R.~W.
\newblock 1962.
\newblock Algorithm 97: Shortest path.
\newblock {\em Commun. ACM} 5(6):345--.

\bibitem[\protect\citeauthoryear{Gibney}{2016}]{gibney2016google}
Gibney, E.
\newblock 2016.
\newblock Google ai algorithm masters ancient game of go.
\newblock {\em Nature News} 529(7587):445.

\bibitem[\protect\citeauthoryear{Gupta \bgroup et al\mbox.\egroup
  }{2017}]{gupta2017cognitive}
Gupta, S.; Davidson, J.; Levine, S.; Sukthankar, R.; and Malik, J.
\newblock 2017.
\newblock Cognitive mapping and planning for visual navigation.
\newblock In {\em The IEEE Conference on Computer Vision and Pattern
  Recognition (CVPR)}.

\bibitem[\protect\citeauthoryear{Johnson}{1977}]{johnson1977efficient}
Johnson, D.~B.
\newblock 1977.
\newblock Efficient algorithms for shortest paths in sparse networks.
\newblock {\em J. ACM} 24(1):1--13.

\bibitem[\protect\citeauthoryear{Kaelbling}{1993}]{kaelbling1993learning}
Kaelbling, L.~P.
\newblock 1993.
\newblock Learning to achieve goals.
\newblock In {\em IJCAI},  1094--1099.
\newblock Citeseer.

\bibitem[\protect\citeauthoryear{Kurutach \bgroup et al\mbox.\egroup
  }{2018}]{kurutach2018model}
Kurutach, T.; Clavera, I.; Duan, Y.; Tamar, A.; and Abbeel, P.
\newblock 2018.
\newblock Model-ensemble trust-region policy optimization.
\newblock {\em arXiv preprint arXiv:1802.10592}.

\bibitem[\protect\citeauthoryear{Lakshminarayanan, Pritzel, and
  Blundell}{2017}]{lakshminarayanan2017simple}
Lakshminarayanan, B.; Pritzel, A.; and Blundell, C.
\newblock 2017.
\newblock Simple and scalable predictive uncertainty estimation using deep
  ensembles.
\newblock In {\em Advances in Neural Information Processing Systems},
  6402--6413.

\bibitem[\protect\citeauthoryear{Levine \bgroup et al\mbox.\egroup
  }{2018}]{levine2018learning}
Levine, S.; Pastor, P.; Krizhevsky, A.; Ibarz, J.; and Quillen, D.
\newblock 2018.
\newblock Learning hand-eye coordination for robotic grasping with deep
  learning and large-scale data collection.
\newblock {\em The International Journal of Robotics Research}
  37(4-5):421--436.

\bibitem[\protect\citeauthoryear{Mirowski \bgroup et al\mbox.\egroup
  }{2016}]{mirowski2016learning}
Mirowski, P.; Pascanu, R.; Viola, F.; Soyer, H.; Ballard, A.~J.; Banino, A.;
  Denil, M.; Goroshin, R.; Sifre, L.; Kavukcuoglu, K.; et~al.
\newblock 2016.
\newblock Learning to navigate in complex environments.
\newblock {\em arXiv preprint arXiv:1611.03673}.

\bibitem[\protect\citeauthoryear{Mirowski \bgroup et al\mbox.\egroup
  }{2017}]{MiPaViICLR2017}
Mirowski, P.; Pascanu, R.; Viola, F.; Soyer, H.; Ballard, A.~J.; Banino, A.;
  Denil, M.; Goroshin, R.; Sifre, L.; Kavukcuoglu, K.; Kumaran, D.; and
  Hadsell, R.
\newblock 2017.
\newblock Learning to navigate in complex environments.
\newblock In {\em Proceedings of the International Conference on Learning
  Representations (ICLR)}.

\bibitem[\protect\citeauthoryear{Mirowski \bgroup et al\mbox.\egroup
  }{2018}]{mirowski2018learning}
Mirowski, P.; Grimes, M.~K.; Malinowski, M.; Hermann, K.~M.; Anderson, K.;
  Teplyashin, D.; Simonyan, K.; Kavukcuoglu, K.; Zisserman, A.; and Hadsell, R.
\newblock 2018.
\newblock Learning to navigate in cities without a map.
\newblock {\em arXiv preprint arXiv:1804.00168}.

\bibitem[\protect\citeauthoryear{Mnih \bgroup et al\mbox.\egroup
  }{2015}]{MnKaSiNATURE2015}
Mnih, V.; Kavukcuoglu, K.; Silver, D.; Rusu, A.~A.; Veness, J.; Bellemare,
  M.~G.; Graves, A.; Riedmiller, M.; Fidjeland, A.~K.; Ostrovski, G.; et~al.
\newblock 2015.
\newblock Human-level control through deep reinforcement learning.
\newblock {\em Nature} 518(7540):529--533.

\bibitem[\protect\citeauthoryear{Oh \bgroup et al\mbox.\egroup
  }{2016}]{OhChSiICML2016}
Oh, J.; Chockalingam, V.; Singh, S.; and Lee, H.
\newblock 2016.
\newblock Control of memory, active perception, and action in minecraft.
\newblock In {\em International Conference on Machine Learning}.

\bibitem[\protect\citeauthoryear{Parisotto and
  Salakhutdinov}{2017}]{parisotto2017neural}
Parisotto, E., and Salakhutdinov, R.
\newblock 2017.
\newblock Neural map: Structured memory for deep reinforcement learning.
\newblock {\em arXiv preprint arXiv:1702.08360}.

\bibitem[\protect\citeauthoryear{Pong \bgroup et al\mbox.\egroup
  }{2018}]{pong2018temporal}
Pong, V.; Gu, S.; Dalal, M.; and Levine, S.
\newblock 2018.
\newblock Temporal difference models: Model-free deep rl for model-based
  control.
\newblock {\em arXiv preprint arXiv:1802.09081}.

\bibitem[\protect\citeauthoryear{Quillen \bgroup et al\mbox.\egroup
  }{2018}]{quillen2018deep}
Quillen, D.; Jang, E.; Nachum, O.; Finn, C.; Ibarz, J.; and Levine, S.
\newblock 2018.
\newblock Deep reinforcement learning for vision-based robotic grasping: A
  simulated comparative evaluation of off-policy methods.
\newblock {\em arXiv preprint arXiv:1802.10264}.

\bibitem[\protect\citeauthoryear{Savinov, Dosovitskiy, and
  Koltun}{2018}]{savinov2018semi}
Savinov, N.; Dosovitskiy, A.; and Koltun, V.
\newblock 2018.
\newblock Semi-parametric topological memory for navigation.
\newblock {\em arXiv preprint arXiv:1803.00653}.

\bibitem[\protect\citeauthoryear{Schaul \bgroup et al\mbox.\egroup
  }{2015}]{schaul2015universal}
Schaul, T.; Horgan, D.; Gregor, K.; and Silver, D.
\newblock 2015.
\newblock Universal value function approximators.
\newblock In {\em International Conference on Machine Learning},  1312--1320.

\bibitem[\protect\citeauthoryear{Sutton and Barto}{1998}]{SuBaBOOK1998}
Sutton, R.~S., and Barto, A.~G.
\newblock 1998.
\newblock {\em Reinforcement learning: An introduction}, volume~1.
\newblock MIT press Cambridge.

\bibitem[\protect\citeauthoryear{Watkins and
  Dayan}{1992}]{watkins1992qlearning}
Watkins, C.~J., and Dayan, P.
\newblock 1992.
\newblock Q-learning.
\newblock {\em Machine learning} 8(3-4):279--292.

\bibitem[\protect\citeauthoryear{Zhang \bgroup et al\mbox.\egroup
  }{2018}]{zhang2018solar}
Zhang, M.; Vikram, S.; Smith, L.; Abbeel, P.; Johnson, M.~J.; and Levine, S.
\newblock 2018.
\newblock Solar: Deep structured latent representations for model-based
  reinforcement learning.
\newblock {\em arXiv preprint arXiv:1808.09105}.

\bibitem[\protect\citeauthoryear{Zhu \bgroup et al\mbox.\egroup
  }{2017}]{zhu2017target}
Zhu, Y.; Mottaghi, R.; Kolve, E.; Lim, J.~J.; Gupta, A.; Fei-Fei, L.; and
  Farhadi, A.
\newblock 2017.
\newblock Target-driven visual navigation in indoor scenes using deep
  reinforcement learning.
\newblock In {\em Robotics and Automation (ICRA), 2017 IEEE International
  Conference on},  3357--3364.
\newblock IEEE.

\end{thebibliography}
\bibliographystyle{aaai}
\end{document}